\documentclass{article}

\usepackage[accepted]{icml2021_oppo}

\usepackage[utf8]{inputenc} %
\usepackage[T1]{fontenc}    %
\usepackage{hyperref}       %
\usepackage{url}            %
\usepackage{booktabs}       %
\usepackage{amsfonts}       %
\usepackage{nicefrac}       %
\usepackage{microtype}      %
\usepackage{xcolor}         %
\usepackage{amsmath}
\usepackage{cleveref}       %
\usepackage{comment}
\usepackage[colorinlistoftodos]{todonotes}
\usepackage{graphics}
\setuptodonotes{fancyline, color=orange!30}
\usepackage{multirow}
\usepackage[framemethod=default]{mdframed}
\usepackage{enumitem}%

\newcommand{\norm}[2]{\left\lVert #1 \right\rVert_{#2}}

\global\mdfdefinestyle{exampledefault}{%
linecolor=lightgray,linewidth=1pt,%
leftmargin=1cm,rightmargin=1cm,
}

\icmltitlerunning{Exploring Low Rank Training of Deep Neural Networks}

\begin{document}

\twocolumn[
\icmltitle{Exploring Low Rank Training of Deep Neural Networks}

\icmlsetsymbol{equal}{*}

\begin{icmlauthorlist}
\icmlauthor{Siddhartha Rao Kamalakara}{equal,to,goo}
\icmlauthor{Acyr Locatelli}{equal,goo}
\icmlauthor{Bharat Venkitesh}{equal,to}
\icmlauthor{Jimmy Ba}{edto}
\icmlauthor{Yarin Gal}{ed}
\icmlauthor{Aidan N. Gomez}{to,goo,ed}
\end{icmlauthorlist}

\icmlaffiliation{to}{Cohere, Inc., Toronto}
\icmlaffiliation{goo}{FOR.ai}
\icmlaffiliation{ed}{Department of Computer Science, University of Oxford, United Kingdom}
\icmlaffiliation{edto}{Department of Computer Science, University of Toronto, Toronto, Canada}

\icmlcorrespondingauthor{Siddhartha Rao Kamalakara}{sid@cohere.ai}

\icmlkeywords{Machine Learning, ICML}

\vskip 0.3in
]
\printAffiliationsAndNotice{\icmlEqualContribution}

\begin{abstract}

Training deep neural networks in low rank, i.e. with factorised layers, is of particular interest to the community: it offers efficiency over unfactorised training in terms of both memory consumption and training time. Prior work has focused on low rank approximations of pre-trained networks and training in low rank space with additional objectives, offering various ad hoc explanations for chosen practice. We analyse techniques that work well in practice, and through extensive ablations on models such as GPT2 we provide evidence falsifying common beliefs in the field, hinting in the process at exciting research opportunities that still need answering.

\end{abstract}

\section{Introduction}

Recent developments in training very large vision and language models \cite{gpt3,switchtransformer, ViT} have led to an increasing need for efficient training paradigms. Low rank matrix factorisation of layers in a deep neural network can offer significant training speedups (up to 2x) and consumes less memory when compared to its unfactorised counterpart.
While matrix factorisation has been studied extensively in the context of linear networks and their applications to matrix sensing and matrix completion problems, the effects of factorised layers on optimisation are non-trivial. Hence, prior work in this space predominantly focused on low-rank training with additional training objectives, or involved computing factorised approximations \emph{post-training}. There has been limited prior work that focused on training dynamics for low rank deep neural networks.

\textbf{Our contributions:} we examine the recent developments in training low rank networks and question existing beliefs about why techniques like singular value decomposition (SVD) based initialisation and modified $L_2$ regularisation are effective. We start with SVD based initialisation techniques which have been found to be effective in both low-rank and sparsity literature \cite{svdsparsity}. We look to random matrix theory to formally define the distribution of singular values at initialisation in modern neural networks and challenge prior assumptions on their importance. We reveal novel empirical insights about the dynamics of singular values during training of an $L_2$ regularised network and present a hypothesis about why $L_2$ regularisation on the re-composed matrix works better than $L_2$ regularisation on its factors. We also investigate currently held beliefs about effective step size and its correlation with performance. Moreover, we analyse and present experiments with pre-training as a strategy to train better performing low-rank networks. We present a wide array of experiments to support our arguments and to demonstrate the effectiveness and practicality of training low-rank neural networks. 
\begin{figure}[ht]
    \centering
    \includegraphics[width=.9\linewidth]{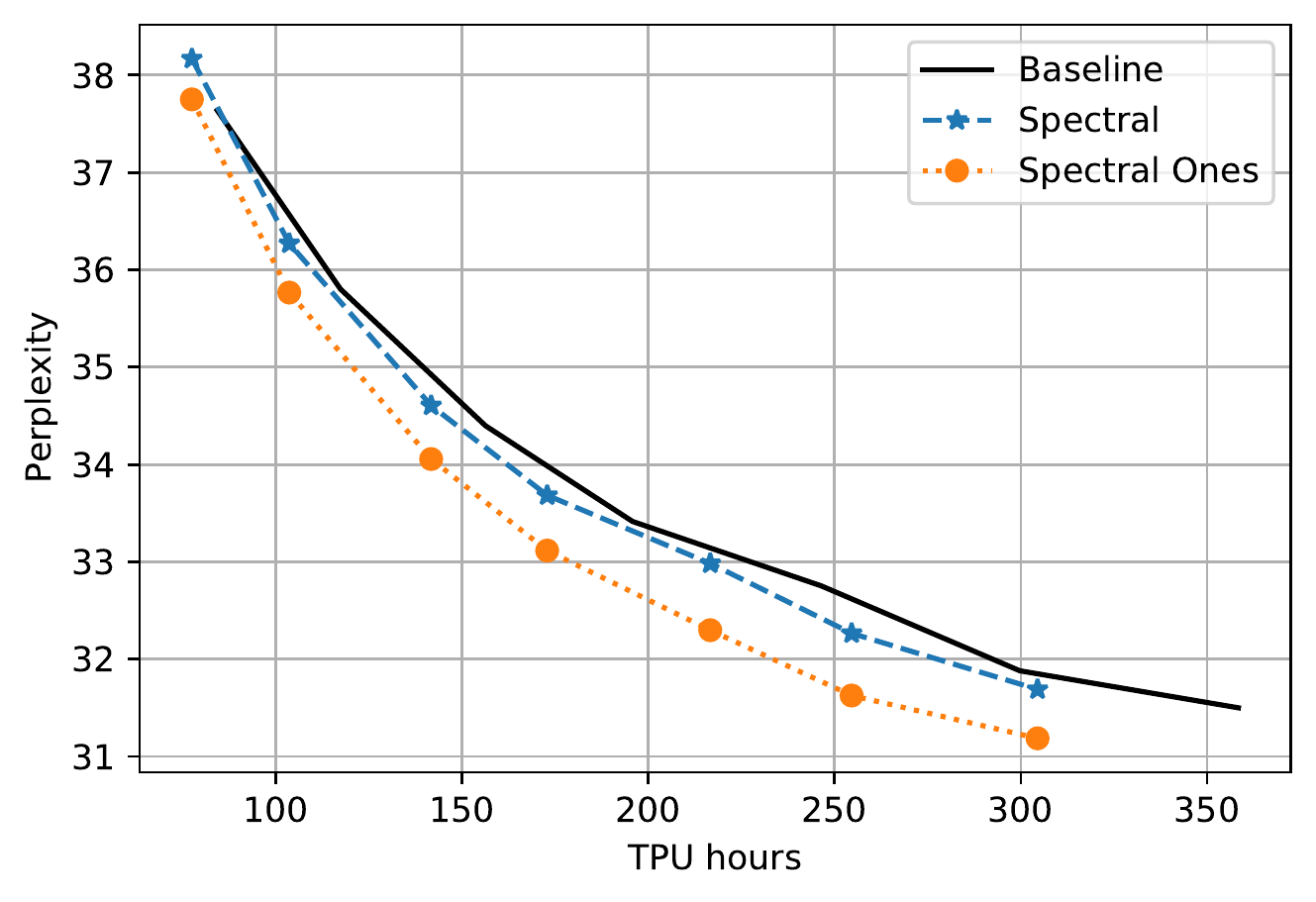}
    \caption{TPU Compute hours vs Performance of GPT-2 on LM1B as the model is scaled up. Each point on the line corresponds to a different model size starting from 1024 hidden dimensions (on the top left) to 2560 (in the bottom right) with increments of 256.}
    \label{computevperf}
\end{figure}
\section{Background}

Most works in the low rank space that focus on efficiency and speedups looked at post-hoc approximation of trained networks. \cite{cvprcompression} took an SVD free approach to reconstruct feature maps by minimising an objective that imposes sparse low rank structure. \cite{filteranddata} also considered a trained network upon which a low rank structure is imposed through filter and data reconstruction objectives. \cite{lowrankcnns} focused on low rank training of CNNs from scratch; they proposed a horizontal and vertical filter decomposition of a convolutional kernel and reproject into orthogonal vectors at every step. One of the reasons why prior work has focused on post-training low rank approximations is that training dynamics of neural networks are poorly understood. Moreover, it has been found that naively training in the low rank space from scratch suffers a gap in performance -- \cref{sec:expts}. To resolve this to an extent, many recent attempts have been made to understand the implicit bias of gradient descent (GD) in matrix factorisation in both linear and non-linear networks. \cite{implicitreg} investigated the behaviour of GD in deep linear networks and found that as the depth of factorisation increases, GD tends to find low rank solutions. They also present evidence for the hypothesis that the language of norms such as nuclear norm, Frobenius norm, etc,  may not be enough to describe the behaviour of GD.  \cite{implicitregmahoney} presented an empirical analysis of commonly used architectures and characterised the dynamics of GD in deep non-linear networks in terms of Empirical Spectral Distributions (ESD) and phases of training. They define a set of rank measures, which we use in our work to analyse low rank training juxtaposed with analysis on unfactored training. \cite{pufferfish} used low rank training with unfactorised pretraining in the context of efficient communication in a distributed setting. \cite{khodak2021initialization} proposed a low rank training procedure by investigating initialisation and regularisation in factorised layers. They analysed SVD based initialisation (Spectral Initialisation) and properties of $L_2$ regularisation which we study independently in our work. They conjecture that there is an interplay between normalisation and weight decay and formalise this behaviour through factorised update equations.
\vspace{-2mm}
\section{Low Rank Training}
In this section, we present the formulation we choose for factorising layers. We discuss and critique the assumptions and conjectures associated with the low rank formulation in the context of SVD initialisation and $L_2$ regularisation.

\subsection{Factorisation}

In all our experiments and analyses, we factorise a weight matrix $W$ at each layer into two components $U$ and $V$ such that $W = UV^\top$. 

We focus on a factorisation depth of 2, taking into consideration memory-speedup tradeoffs: As the depth of factorisation at each layer increases, more activations need to be stored in-memory for backpropagation. A depth of two provides speedups across all our experiments while ensuring minimal activation memory overhead.

Consider the difference between the vanilla gradient descent update (unfactorised) \(W_{t+1} = W_t - \alpha \nabla W\) and the update performed in the factorised setting:
\begin{align*}
W_{t+1} ={}& U_{t+1} V_{t+1}^\top \nonumber \\ 
\end{align*}
\begin{align}
W_{t+1} ={}& (U_t - \alpha \nabla U)(V_t - \alpha \nabla V)^\top \nonumber \\ 
\begin{split}
W_{t+1} ={}& W_t - \alpha \underbrace{(\nabla W_t V_t V_t^\top + U_t U_t^\top \nabla W_t)}_{\nabla_t} \\
{}& + \alpha^2 \nabla W_t W_t \nabla W_t ^ \top
\label{fullupdate}
\end{split}
\end{align}

\cite{khodak2021initialization} extend the update equation above to normalised layers. Most modern architectures rely on normalisation layers to train networks that generalise well. This includes batch normalisation \cite{batchnorm} in ResNets and layer normalisation \cite{layernorm} in Transformers. We refer the reader to \cite{khodak2021initialization} for a more detailed discussion on the type and role of normalisation in factorised layers and use their formulation of the normalised update equation, which is given by
\begin{align}
\begin{split}
\hat{w}_{t+1} ={}& \hat{w_t} - \frac{\alpha}{\norm{W}{F}^2} (I_{mn} - \hat{w_t} \hat{w_t}^\top) \text{vec}(\hat{\nabla}_t) \\
{}& + \mathcal{O}(\alpha^2)
\end{split}
\label{normalizedupdate}
\end{align}
where $\hat{\nabla}_t$ is $\nabla_t$ with gradients taken with respect to the normalised weight matrix $\hat{W} = \frac{W}{\norm{W}{F}}$ and $\hat{w}=\text{vec}(\hat{W})$.

We see that gradient descent in the factorised setting does not perfectly align with the vanilla gradient descent update. In the subsequent sections, we empirically explore and work to overcome the implicit biases of this factorised update so that we can make low rank training an effective and efficient training method.

\subsubsection{Fully connected layer}\label{fcfactor}

Let $W \in \mathbb{R}^{m\times n}$ be the weight matrix of a fully-connected layer. We factorise $W$ as $W = UV^T$ with $U \in R^{m\times r}$ and $V^T \in R^{r\times n}$, where $0 < r \leq \min(m, n)$. At inference, when $r< \frac{m\times n}{m+n}$, factorising the fully connected weight matrix leads to a reduced memory footprint as well as floating point operations (flops) from $\mathcal{O}(mn)$ to $\mathcal{O}(mr + rn)$. For training, the memory requirements change from $\mathcal{O}(mn + n)$ to $\mathcal{O}(mr + rn + n+ r)$ as we need to store the intermediate activations for backpropagation.

\subsubsection{Convolutional layer} \label{convfactor}

We factorise convolution kernels in a way that supports rewriting the single convolution as two convolutions. We choose to factorise the convolutional kernel $W \in R^{h \times w \times c_{in} \times c_{out}}$ as $W = UV^T$ with $U \in R^{h \times w \times c_{in} \times r}$ and $V^T \in R^{1 \times 1 \times r \times c_{out}}$ where $h, w$ represent the kernel height and width respectively, $c_{in}$ and $c_{out}$ represent the number of input and output channels respectively and $r$ represents the rank of the decomposition. In the low-rank decomposition, $r \leq \min(h \times w \times c_{in}, c_{out})$. This leads to a reduction in flops from $\mathcal{O}(hwc_{in}c_{out})$ to $\mathcal{O}(hwc_{in}r+rc_{out}).$

\subsection{Spectral Initialisation}
 \cite{khodak2021initialization} investigated the usefulness of spectral initialisation in low rank formulations of deep learning architectures and proposed a few hypotheses for why it works. We use the same truncated SVD initialisation scheme, which is defined as follows:
\begin{align}
\text{SVD}_r(W) &= \hat{U}_{:r} \Sigma_r \hat{V}^\top_{:r} \hspace{0.5mm}\text{,} \label{eqn:svd} \\
U &= \hat{U}_{:r} \sqrt{\Sigma_r}\text{,} \nonumber \\
V &= \hat{V}_{:r} \sqrt{\Sigma_r}\text{,} \nonumber
\end{align}
where $W$ is a matrix of shape $N \times M$, $U$ of shape $N \times r$, $V$ of shape $M \times r$, \(\Sigma\) is the diagonal matrix of singular values and $r$ is the rank we choose for the factorisation. We note that $U$ and $V$ are rectangular matrices unless specified otherwise.

\cite{khodak2021initialization} analysed SVD based initialisation in the context of the update Equation \ref{fullupdate} and provide two hypotheses for why this technique works, both of which we disprove. 

\begin{itemize}
    \item $U_0 U_0^\top = V_0 V_0^\top = \Sigma_r$. \\ \\
In the low rank context, $U$ and $V$ are rectangular matrices obtained from truncated SVD which makes $U$ and $V$ column-wise orthogonal matrices. Therefore, $UU^\top$ and $VV^\top$ \emph{cannot} be equal to $\Sigma_{r}$ and $\nabla W_t V_t V_t^\top + U_t U_t^\top \nabla W_t$ terms in the Equation \ref{fullupdate} cannot be simplified.
    \item The singular values of a Gaussian ensemble of scale $\frac{1}{\sqrt{n}}$ are roughly distributed around 1. \\ \\
We look to Marchenko-Pastur theory (described in Appendix \ref{mptheory}) to understand the distribution of singular values of a Gaussian ensemble matrix of size $N \times M$, which states that the distribution of singular values is dependent on the scale of the random initialisation $\sigma^2$ and the size ratio $\frac{N}{M}$ of the layer.
\end{itemize}

We believe that spectral initialisation works for reasons other than the ones stated in prior work. In Section \ref{expt:init}, we present an ablation experiment that hints at why this initialisation scheme performs better.

\subsection{$L_2$ Regularisation}
Many architectures rely on $L_2$ regularisation for better generalisation. The straightforward approach to impose $L_2$ regularisation in a factorised network is to apply the Frobenius norm penalty to the factors $U$ and $V$ -- that is, $\frac{\lambda}{2}(\norm{U}{F}^2+\norm{V}{F}^2)$.  \cite{nuclearnormlemma} showed that this penalty actually minimises the nuclear norm of the recomposed matrix $UV^\top$.

To address this, \cite{khodak2021initialization} propose penalising the Frobenius norm of the recomposed matrix $UV^\top$, which they refer to as, Frobenius decay. They argue that Frobenius decay helps in keeping the effective step size high through out training where effective step size is the term $\frac{\eta}{\norm{W}{F}^2}$ in Equation \ref{normalizedupdate}. We show, through an ablations study, that effective step size is an inadequate argument to justify the effectiveness of Frobenius decay over $L_2$ regularization. We point out that the dynamics of low-rank training with $L_2$ regularisation cannot be understood by only considering the normalised update Equation \ref{normalizedupdate}. This ignores the $ \eta \lambda \approx \mathcal{O}(\eta^2)$ terms arising from Frobenius norm penalty which have a non-trivial impact on the optimisation.
We find that the effectiveness of Frobenius decay over $L_2$ regularisation can be better explained by examining the effective rank of the network. We use the rank measure proposed in  \cite{implicitregmahoney} which defines effective rank of a matrix $W$ to be:
$$\frac{\norm{W}{*}}{\norm{W}{\text{op}}}.$$
That is, the ratio between nuclear norm and the operator norm. In our case, we are interested in the effective rank of $UV^{\top}$

\subsection{Pre-training} \label{pretraining}
The initial stages of training are widely believed to be important for good performance in neural networks \cite{criticalperiod} \cite{lateresetting}. This motivates us to explore training for a fraction of the total training steps in the unfactorised space before switching to low rank substitutions of these unfactorised layers. We apply the truncated SVD scheme descibed in Equation \ref{eqn:svd} to the partially trained weights to obtain the factors of the layer. Section \ref{expt:pretrain} describes the impact of pre-training on performance across our vision and language experiments and analyses the nature of the solutions found with pre-training when compared to solutions found by low rank networks trained from scratch \cite{interpolation} \cite{linearmodeconn}.

\vspace{-2mm}
\section{Experiments and Results}
\label{sec:expts}
We conduct extensive experiments on both vision and language models. For vision models, we use a Wide-ResNet-28 \cite{wideresnet} on CIFAR-100 and a ResNet-50 \cite{resnet} on the ImageNet dataset. For the language modelling task, we conduct experiments on one million word benchmark dataset (LM1B) \cite{lm1b} and use the GPT-2 \cite{gpt2} architecture. Details on our complete experimental setup can be found in Appendix \ref{appendix:exp_details}. In the following sections, we compare different initialisation schemes and study the effects of $L_2$ regularisation and Frobenius decay. Finally, we demonstrate the effectiveness of \textemdash{}   and analyse the nature of solutions found by \textemdash{} pre-training.
\vspace{-2mm}
\subsection{Initialisation} \label{expt:init}
We show that spectral initialisation offers equivalent performance when compared to traditional initialisation schemes. Then, we show empirically that the singular values do not play a major role in improving performance and that it is the direction of the singular vectors that matters. This finding is in contrast with prior beliefs \cite{khodak2021initialization} about the role of singular values in retaining the scale of initialisation.  We establish this by setting the singular values to ones in Equation \ref{eqn:svd}.
Tables \ref{wrn_init}, \ref{imagenet_init}, \ref{transformer_init} compare the results across initialisation schemes on CIFAR100, ImageNet and LM1B respectively. We observe that spectral ones leads to a better accuracy on CIFAR-100, lower perplexity on LM1B and a commensurate performance on ImageNet.

\subsection{$L_2$ Regularisation}

We investigate the effective step size hypothesis by training two networks, one with learning rate $\eta$ and the other with $\frac{\eta}{2}$. So, the effective step size of these networks is $\frac{\eta}{\norm{W}{F}^2}$ and $\frac{\eta}{2\norm{W}{F}^2}$ respectively, based on Equation \ref{normalizedupdate}. If the hypothesis that a higher effective step size leads to better performance were true, we should see that halving the effective step size should lead to a lower performance but we find that $\frac{\eta}{2}$ leads to models that are atleast as good as models trained with learning rate $\eta$.

Tables \ref{wrn_l2}, \ref{imagenet_l2} and \ref{transformer_l2} compare the impact of effective step size on performance across CIFAR-100, ImageNet and LM1B respectively. Analysing the evolution of singular values in networks trained with $L_2$ regularisation and Frobenius decay revealed that singular values are disproportionately affected in the case of $L_2$ regularisation. We observe a "rich get richer, poor get poorer" phenomenon in $L_2$ regularised networks which causes the effective rank $\frac{\norm{UV^\top}{*}}{\norm{UV^\top}{op}}$ of the network to drop because of the disproportionate increase in the operator norm of each layer. We report the averaged (across layers) effective rank at the end of training for our experiments in Table \ref{erank_table}.
\begin{table}[H]
    \centering
    \begin{tabular}{|c|c|c|c|}
    \hline
         Model& Dataset& Frobenius decay & $L_2$   \\ \hline
         WRN& CIFAR-100& 39.87 & 16.4\\ \hline
         ResNet-50 & ImageNet & 68.72 & 58.00 \\ \hline
         Transformer & LM1B & 206.93 & 205.70\\ \hline
    \end{tabular}
    \caption{Effective rank measures for different models}
    \label{erank_table}
\end{table}

\subsection{Pre-training} \label{expt:pretrain}

\begin{figure}[t]
    \centering
    \includegraphics[width=\linewidth]{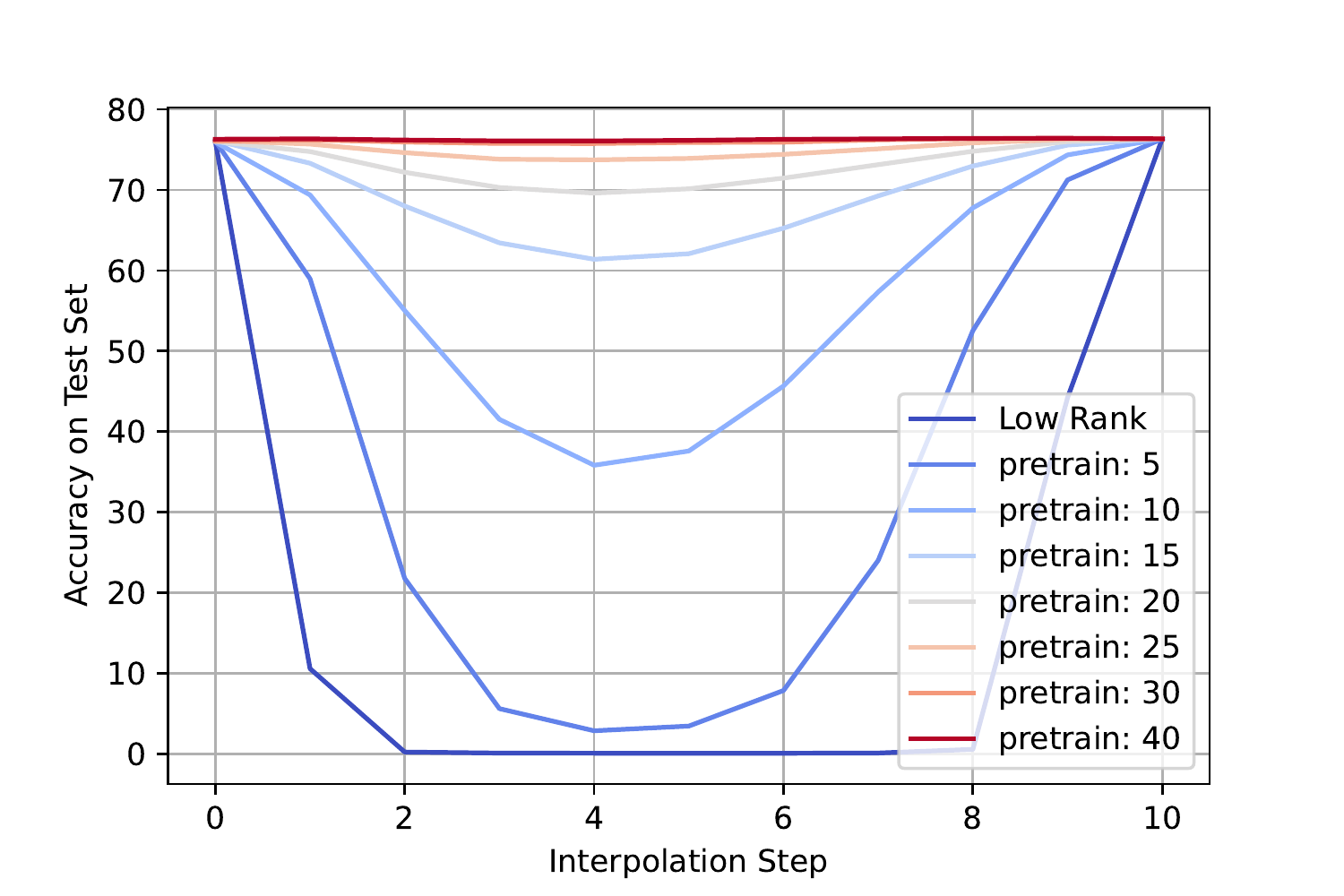}
    \caption{Comparison of interpolation of low rank and pre-trained networks for ResNet-50 on ImageNet with a rank of 50 \%.}
    \label{interpolation_imagenet}
\end{figure}

We investigate pre-training networks for a fraction of the total training steps and observe that this leads to a significantly improved performance in our language model experiments as shown in Figures \ref{computevperf} and \ref{paramsvperf} when we scale up the model. We pre-train in the unfactorised space for 40,000 steps and continue training in the factorised space for 200,000 steps. We combine pre-training with the techniques aforementioned \textit{viz} Frobenius decay and resuming with decompositions obtained from Spectral and Spectral ones as described in \ref{pretraining}. We find that pre-training does not offer improved performance compared to low-rank network trained from scratch in our vision experiments as shown in Tables \ref{pretrain:cifar100} and \ref{pretrain:imagenet}. Furthermore, we notice that the solutions found with pre-training are closer in the parameter space to their corresponding baseline (unfactorised) models. We demonstrate this by performing linear interpolation, shown in Figures \ref{interpolation_imagenet}, \ref{interpolation_wrn} and \ref{interpolationLM}, between pre-training and baseline weights by using the following equation:
$\theta = (1-t) \theta_b +t\theta_l$ for $t \in [0.0, 1.0]$ with increments of 0.1 where $t$ is the interpolation coefficient, $\theta_b$ is the parameter from the baseline model and $\theta_l$ is the parameter from the low rank model with pre-training.

\section{Conclusion}

We demonstrated empirically that Spectral initialisation and $L_2$ regularisation on $UV^\top$ improve low-rank training but are poorly understood. We presented singular value analyses and ablation studies that act as counter-examples to prior beliefs about why these techniques work. We hope to put forth the theoretical reasons behind the effectiveness of these techniques in a future work. Additionally, we demonstrated pretraining as an effective strategy to improve low-rank performance and presented insights on the nature of solutions found by networks with pretraining. 
\bibliography{references}

\begin{thebibliography}{22}
\providecommand{\natexlab}[1]{#1}
\providecommand{\url}[1]{\texttt{#1}}
\expandafter\ifx\csname urlstyle\endcsname\relax
  \providecommand{\doi}[1]{doi: #1}\else
  \providecommand{\doi}{doi: \begingroup \urlstyle{rm}\Url}\fi

\bibitem[Achille et~al.(2017)Achille, Rovere, and Soatto]{criticalperiod}
Achille, A., Rovere, M., and Soatto, S.
\newblock Critical learning periods in deep neural networks.
\newblock \emph{CoRR}, abs/1711.08856, 2017.
\newblock URL \url{http://arxiv.org/abs/1711.08856}.

\bibitem[Arora et~al.(2019)Arora, Cohen, Hu, and Luo]{implicitreg}
Arora, S., Cohen, N., Hu, W., and Luo, Y.
\newblock Implicit regularization in deep matrix factorization, 2019.

\bibitem[Ba et~al.(2016)Ba, Kiros, and Hinton]{layernorm}
Ba, J.~L., Kiros, J.~R., and Hinton, G.~E.
\newblock Layer normalization, 2016.

\bibitem[Brown et~al.(2020)Brown, Mann, Ryder, Subbiah, Kaplan, Dhariwal,
  Neelakantan, Shyam, Sastry, Askell, Agarwal, Herbert-Voss, Krueger, Henighan,
  Child, Ramesh, Ziegler, Wu, Winter, Hesse, Chen, Sigler, Litwin, Gray, Chess,
  Clark, Berner, McCandlish, Radford, Sutskever, and Amodei]{gpt3}
Brown, T.~B., Mann, B., Ryder, N., Subbiah, M., Kaplan, J., Dhariwal, P.,
  Neelakantan, A., Shyam, P., Sastry, G., Askell, A., Agarwal, S.,
  Herbert-Voss, A., Krueger, G., Henighan, T., Child, R., Ramesh, A., Ziegler,
  D.~M., Wu, J., Winter, C., Hesse, C., Chen, M., Sigler, E., Litwin, M., Gray,
  S., Chess, B., Clark, J., Berner, C., McCandlish, S., Radford, A., Sutskever,
  I., and Amodei, D.
\newblock Language models are few-shot learners, 2020.

\bibitem[Chelba et~al.(2013)Chelba, Mikolov, Schuster, Ge, Brants, and
  Koehn]{lm1b}
Chelba, C., Mikolov, T., Schuster, M., Ge, Q., Brants, T., and Koehn, P.
\newblock One billion word benchmark for measuring progress in statistical
  language modeling.
\newblock \emph{CoRR}, abs/1312.3005, 2013.
\newblock URL \url{http://arxiv.org/abs/1312.3005}.

\bibitem[Dosovitskiy et~al.(2020)Dosovitskiy, Beyer, Kolesnikov, Weissenborn,
  Zhai, Unterthiner, Dehghani, Minderer, Heigold, Gelly, Uszkoreit, and
  Houlsby]{ViT}
Dosovitskiy, A., Beyer, L., Kolesnikov, A., Weissenborn, D., Zhai, X.,
  Unterthiner, T., Dehghani, M., Minderer, M., Heigold, G., Gelly, S.,
  Uszkoreit, J., and Houlsby, N.
\newblock An image is worth 16x16 words: Transformers for image recognition at
  scale.
\newblock \emph{CoRR}, abs/2010.11929, 2020.
\newblock URL \url{https://arxiv.org/abs/2010.11929}.

\bibitem[Evci et~al.(2019)Evci, Pedregosa, Gomez, and Elsen]{interpolation}
Evci, U., Pedregosa, F., Gomez, A.~N., and Elsen, E.
\newblock The difficulty of training sparse neural networks.
\newblock \emph{CoRR}, abs/1906.10732, 2019.
\newblock URL \url{http://arxiv.org/abs/1906.10732}.

\bibitem[Fedus et~al.(2021)Fedus, Zoph, and Shazeer]{switchtransformer}
Fedus, W., Zoph, B., and Shazeer, N.
\newblock Switch transformers: Scaling to trillion parameter models with simple
  and efficient sparsity.
\newblock \emph{CoRR}, abs/2101.03961, 2021.
\newblock URL \url{https://arxiv.org/abs/2101.03961}.

\bibitem[Frankle et~al.(2019{\natexlab{a}})Frankle, Dziugaite, Roy, and
  Carbin]{lateresetting}
Frankle, J., Dziugaite, G.~K., Roy, D.~M., and Carbin, M.
\newblock The lottery ticket hypothesis at scale.
\newblock \emph{CoRR}, abs/1903.01611, 2019{\natexlab{a}}.
\newblock URL \url{http://arxiv.org/abs/1903.01611}.

\bibitem[Frankle et~al.(2019{\natexlab{b}})Frankle, Dziugaite, Roy, and
  Carbin]{linearmodeconn}
Frankle, J., Dziugaite, G.~K., Roy, D.~M., and Carbin, M.
\newblock Linear mode connectivity and the lottery ticket hypothesis.
\newblock \emph{CoRR}, abs/1912.05671, 2019{\natexlab{b}}.
\newblock URL \url{http://arxiv.org/abs/1912.05671}.

\bibitem[He et~al.(2015)He, Zhang, Ren, and Sun]{resnet}
He, K., Zhang, X., Ren, S., and Sun, J.
\newblock Deep residual learning for image recognition.
\newblock \emph{CoRR}, abs/1512.03385, 2015.
\newblock URL \url{http://arxiv.org/abs/1512.03385}.

\bibitem[Ioffe \& Szegedy(2015)Ioffe and Szegedy]{batchnorm}
Ioffe, S. and Szegedy, C.
\newblock Batch normalization: Accelerating deep network training by reducing
  internal covariate shift, 2015.

\bibitem[Jaderberg et~al.(2014)Jaderberg, Vedaldi, and
  Zisserman]{filteranddata}
Jaderberg, M., Vedaldi, A., and Zisserman, A.
\newblock Speeding up convolutional neural networks with low rank expansions,
  2014.

\bibitem[Khodak et~al.(2021)Khodak, Tenenholtz, Mackey, and
  Fusi]{khodak2021initialization}
Khodak, M., Tenenholtz, N.~A., Mackey, L., and Fusi, N.
\newblock Initialization and regularization of factorized neural layers.
\newblock In \emph{International Conference on Learning Representations}, 2021.
\newblock URL \url{https://openreview.net/forum?id=KTlJT1nof6d}.

\bibitem[Lee et~al.(2019)Lee, Ajanthan, Gould, and Torr]{svdsparsity}
Lee, N., Ajanthan, T., Gould, S., and Torr, P. H.~S.
\newblock A signal propagation perspective for pruning neural networks at
  initialization.
\newblock \emph{CoRR}, abs/1906.06307, 2019.
\newblock URL \url{http://arxiv.org/abs/1906.06307}.

\bibitem[Martin \& Mahoney(2018)Martin and Mahoney]{implicitregmahoney}
Martin, C.~H. and Mahoney, M.~W.
\newblock Implicit self-regularization in deep neural networks: Evidence from
  random matrix theory and implications for learning, 2018.

\bibitem[Radford et~al.(2019)Radford, Wu, Child, Luan, Amodei, and
  Sutskever]{gpt2}
Radford, A., Wu, J., Child, R., Luan, D., Amodei, D., and Sutskever, I.
\newblock Language models are unsupervised multitask learners.
\newblock 2019.

\bibitem[Srebro \& Shraibman(2005)Srebro and Shraibman]{nuclearnormlemma}
Srebro, N. and Shraibman, A.
\newblock Rank, trace-norm and max-norm.
\newblock In Auer, P. and Meir, R. (eds.), \emph{Learning Theory}, pp.\
  545--560, Berlin, Heidelberg, 2005. Springer Berlin Heidelberg.
\newblock ISBN 978-3-540-31892-7.

\bibitem[Tai et~al.(2016)Tai, Xiao, Zhang, Wang, and E]{lowrankcnns}
Tai, C., Xiao, T., Zhang, Y., Wang, X., and E, W.
\newblock Convolutional neural networks with low-rank regularization, 2016.

\bibitem[Wang et~al.(2021)Wang, Agarwal, and Papailiopoulos]{pufferfish}
Wang, H., Agarwal, S., and Papailiopoulos, D.
\newblock Pufferfish: Communication-efficient models at no extra cost, 2021.

\bibitem[Yu et~al.(2017)Yu, Liu, Wang, and Tao]{cvprcompression}
Yu, X., Liu, T., Wang, X., and Tao, D.
\newblock On compressing deep models by low rank and sparse decomposition.
\newblock pp.\  67--76, 2017.
\newblock \doi{10.1109/CVPR.2017.15}.

\bibitem[Zagoruyko \& Komodakis(2016)Zagoruyko and Komodakis]{wideresnet}
Zagoruyko, S. and Komodakis, N.
\newblock Wide residual networks.
\newblock \emph{CoRR}, abs/1605.07146, 2016.
\newblock URL \url{http://arxiv.org/abs/1605.07146}.

\end{thebibliography}
\bibliographystyle{icml2021.bst}

\clearpage

\clearpage
\appendix

\section{Appendix}

\subsection{Marchenko-Pastur Theory} \label{mptheory}

Marchenko-Pastur (MP) theory defines the distribution of singular values of Gaussian random matrices in the infinite limit but is applicable to finite matrices with very reasonable error bounds. MP theory defines the distribution as:
\begin{align}
\rho (\lambda) &= 
\begin{cases}
\frac{N}{2 \pi \sigma^2M} \frac{\sqrt{(\lambda^+ - \lambda)(\lambda - \lambda^-)}}{\lambda} \hspace{1.0mm} & \text{if } \lambda \in [\lambda^-, \lambda^+] \\
0 & \text{ otherwise} \\
\end{cases}
\end{align}

\begin{align}
\lambda^\pm &= \sigma^2 \bigg( 1 \pm \sqrt{\frac{M}{N}}\bigg)^2\text{,}
\end{align}

\subsection{Experiment Details}
\label{appendix:exp_details}
For the language modelling task, we conduct our experiments on one million word benchmark dataset (LM1B) \cite{lm1b} and use the following set up: input sequence length is fixed at 256 and 1152 tokens for training and evaluation respectively and the vocab size is limited to 32K subwords and train all the models to 240K steps. We implemented transformer language model on Tensorflow and run all our experiments on cloud TPUs. To have better savings on compute and memory we combine the query, key value generation into one weight matrix. For each transformer layer, we decompose three matrix operations; Q,K,V generation and the two fully connected layers. We skip factorising the output projection layer and the combiner layer that combines the outputs of attention (this is a square matrix and we see memory and compute benefit only for very small ranks). For all transformer runs, we choose a rank of 62.5\% and half its baseline learning rate. For pre-training, we train unfactored for 40K steps then switch to low rank factorised training for the remaining 200K steps and halving the learning rate.

For the image classification task, we conduct experiments with CIFAR-100 and ImageNet. For CIFAR-100 we use the standard training/test split with a simple augmentation scheme -- Random Crop and Horizontal Flips. We train a WideResNet-28 \cite{wideresnet} for 200 epochs with SGD with momentum (0.9) and a batch size of 128. For regularisation, we a weight decay coefficient of 5e-4 and no dropout. For the low rank training runs, we factorised every convolutional layer other than the first according to our factorisation scheme describe above and the chosen rank. For ImageNet experiments, we use a standard ResNet-50 architecture and train on a TPU v2-8 with a per-core batch size of 128 and follow the same hyperparameters and learning rate schedule described in \cite{resnet}.

\subsection{Initialization Results}
\begin{table}[H]
\centering
\begin{tabular}{|c|c|c|}
\hline
\textbf{Rank} & \textbf{Initialisation} & \textbf{Accuracy} \\
\hline
Baseline (N/A) & He & 81.08 \\
\hline

\multirow{3}{*}{0.1} & He & 77.94 \\ 
                    & spectral   & 79.84 \\
                  & spectral ones & 79.07 \\ 
\hline
\multirow{3}{*}{0.2} & He & 80.37 \\
                    & spectral   & 81.35 \\
                  & spectral ones & 81.27 \\ 
\hline
\multirow{3}{*}{0.3} & He & 80.87 \\  
                    & spectral   & 81.53 \\
                    & spectral ones & 81.61 \\ 
\hline
\end{tabular}
\caption{Initialization results of Wide Resnets on Cifar-100}
\label{wrn_init}
\end{table}

\begin{table}[h]
\centering
\begin{tabular}{|c|c|c|c|}
\hline
\textbf{Rank} & \textbf{Initialisation} & \textbf{Top-1} & \textbf{Top-5} \\
\hline
Baseline (N/A) & He & 76.39 & 93.21 \\
\hline
\multirow{3}{*}{0.3} & He  & 75.26 & 92.56 \\ 
                & spectral   & 75.77 & 92.87 \\
                & spectral ones   & 75.71 & 92.82 \\
\hline
\multirow{3}{*}{0.5} & He  & 75.97 & 92.84 \\
                & spectral   & 76.13 & 93.09 \\
                & spectral ones   & 75.98 & 92.97 \\
\hline
\end{tabular}
\caption{Initialization results of ResNet on Image Net}
\label{imagenet_init}
\end{table}

\begin{table}[h]
\centering
\begin{tabular}{|c|c|c|}
\hline
\textbf{Rank} & \textbf{Initialisation} & \textbf{Perplexity} \\
\hline
Baseline (N/A) & He & 37.67\\
\hline
\multirow{3}{*}{0.62} & He   & 39.6\\
                & spectral   & 38.78 \\
                   & spectral ones  & 38.47 \\
\hline
\end{tabular}
\caption{Initialization results of Transformers on LM1B}
\label{transformer_init}
\end{table}

\subsection{Regularization Results}
\begin{table}[h]
\centering
\begin{tabular}{|c|c|c|c|}
\hline
Rank & \textbf{Regularisation} & \textbf{lr scaling} & \textbf{Accuracy} \\
\hline
\hline

\multirow{4}{*}{0.1} &  \multirow{2}{*}{L2} & 0.5 & 73.12 \\
                     &                       & 1.0 & 72.59 \\
\cline{2-4}
                     &  \multirow{2}{*}{Frobenius Decay} & 0.5 & 79.84 \\
                     &                       & 1.0 & 79.79 \\
\hline

\multirow{4}{*}{0.2} &  \multirow{2}{*}{L2} & 0.5 & 78.22 \\
                     &                       & 1.0 & 77.56 \\
\cline{2-4}
                     &  \multirow{2}{*}{Frobenius Decay} & 0.5 & 81.35 \\
                     &                       & 1.0 & 81.61 \\
\hline

\end{tabular}
\caption{Comparison between Frobenius Decay and L2 regularisation on Cifar-100}
\label{wrn_l2}
\end{table}

\begin{table}[t]
\centering
\begin{tabular}{|c|c|c|c|c|}
\hline
\textbf{Rank} & \textbf{Regularization} & \textbf{lr scaling} & \textbf{Top-1} & \textbf{Top-5} \\
\hline
\hline

\multirow{4}{*}{0.3} & \multirow{2}{*}{L2}  & 0.5  & 75.11 & 92.42 \\
                & & 1.0 & 74.9 & 92.24 \\
\cline{2-5}
& \multirow{2}{*}{Frobenius Decay} & 0.5  & 75.22 & 92.49 \\
                   &  & 1.0 & 75.77 & 92.87 \\
\hline
\multirow{4}{*}{0.5} & \multirow{2}{*}{L2} & 0.5  & 75.04 & 92.36 \\
                   & & 1.0 & 74.83 & 92.25 \\
\cline{2-5}
& \multirow{2}{*}{Frobenius Decay} &  0.5  & 75.97 & 92.85 \\
                   & & 1.0 & 76.13 & 93.09 \\
\hline

\end{tabular}
\caption{Comparison between Frobenius Decay and L2 regularisation on Imagenet}
\label{imagenet_l2}
\end{table}

\begin{table}[t]
\centering
\begin{tabular}{|c|c|c|c|}
\hline
Rank & \textbf{Regularisation} & \textbf{lr scaling} & \textbf{Perplexity} \\
\hline
\hline

\multirow{4}{*}{0.62} &  \multirow{2}{*}{L2} & 0.5 & 38.87 \\
                     &                       & 1.0 & 39.01 \\
\cline{2-4}
                     &  \multirow{2}{*}{Frobenius Decay} & 0.5 & 38.78 \\
                     &                       & 1.0 & 39.2 \\
\hline

\end{tabular}
\caption{Comparison between Frobenius Decay and L2 regularisation on LM1B}
\label{transformer_l2}
\end{table}

\subsection{Pre-training Results}
\begin{table}[H]
\centering
\begin{tabular}{|c|c|c|c}
\hline
\textbf{Rank} & \textbf{Pre-training Epochs} &  \textbf{Accuracy}\\
\hline\hline
\multirow{7}{*}{0.2} & 0 & 81.35 \\
                     & 15 & 81.33 \\ 
                     & 30 & 81.56 \\
                     & 40 & 81.53 \\
                     & 50 & 81.39 \\
                     & 75 & 81.53 \\
                     
\hline
\multirow{7}{*}{0.3} & 0 & 81.53 \\
                     & 15 & 81.73 \\ 
                     & 30 & 81.51 \\
                     & 40 & 81.67 \\
                     & 50 & 82.0 \\
                     & 75 & 81.44 \\
\hline
\end{tabular}
\caption{Pre-training results for Wide ResNets on CIFAR-100}
\label{pretrain:cifar100}
\end{table}
\begin{table}[H]
\centering
\begin{tabular}{|c|c|c|c|}
\hline
\textbf{Rank} & \textbf{\# Pretrain epochs} &  \textbf{Top-1} & \textbf{Top-5} \\
\hline\hline
\multirow{9}{*}{0.5} & 5 & 76.07 & 92.88 \\
                     & 10 & 75.96 & 93.04 \\
                     & 15 & 76.12 & 92.96 \\ 
                     & 20 & 76.08 & 92.94 \\
                     & 25 & 76.15 & 93.00 \\
                     & 30 & 76.05 & 92.9 \\
                     & 35 & 76.24 & 93.06 \\
                     & 40 & 76.21 & 93.09 \\
                     & 45 & 76.29 & 93.12 \\
\hline
\end{tabular}
\caption{Pre-training results for ResNet50 on ImageNet}
\label{pretrain:imagenet}
\end{table}

\begin{figure}[H]
    \centering
    \includegraphics[width=\linewidth]{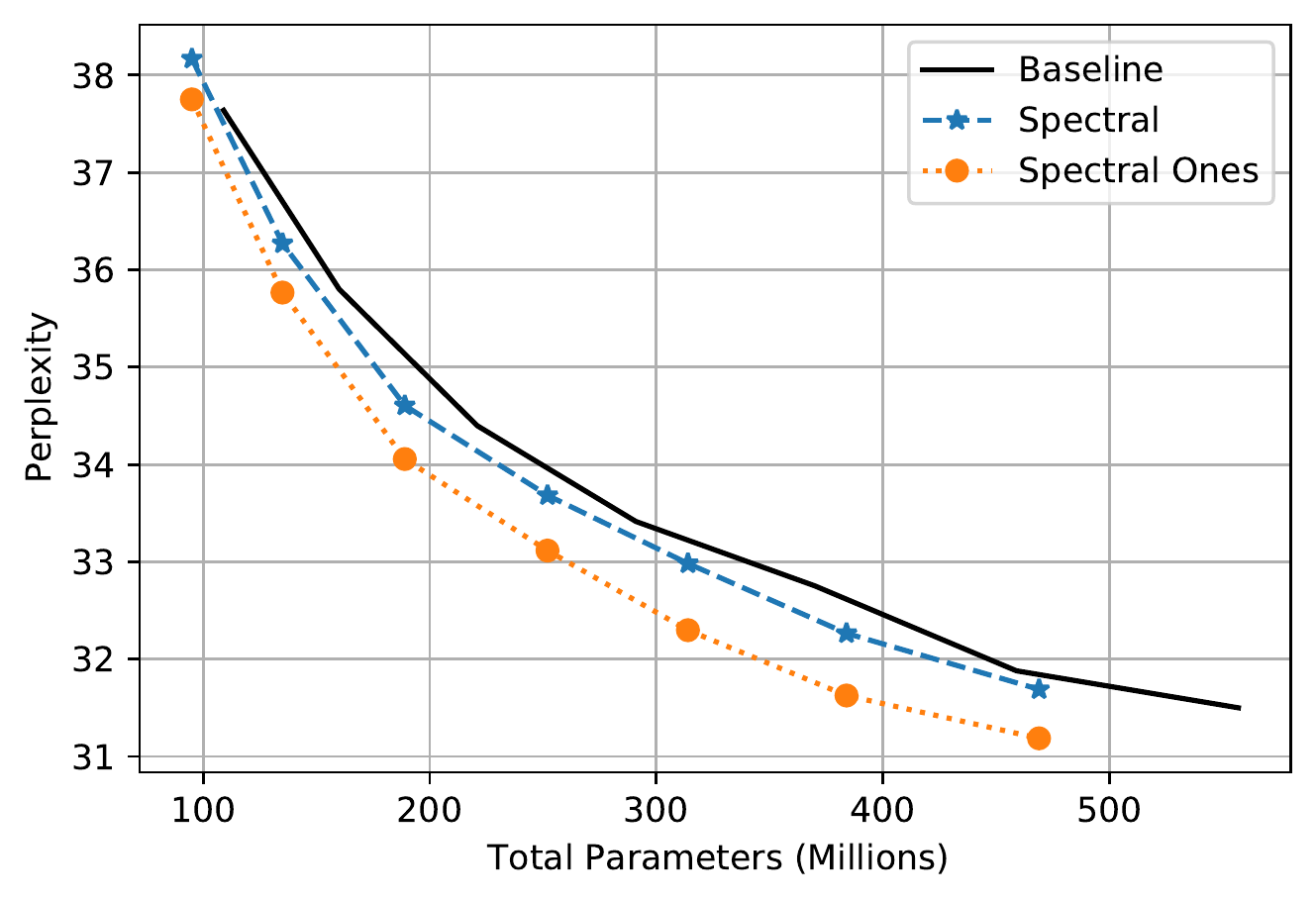}
    \caption{Total parameters vs Performance of GPT-2 on LM1B as the model is scaled up. Each point on the line corresponds to a different model size starting from 1024 hidden dimensions (on the top left) to 2560 (in the bottom right) with increments of 256.}
    \label{paramsvperf}
\end{figure}

\begin{figure}[H]
    \centering
    \includegraphics[width=\linewidth]{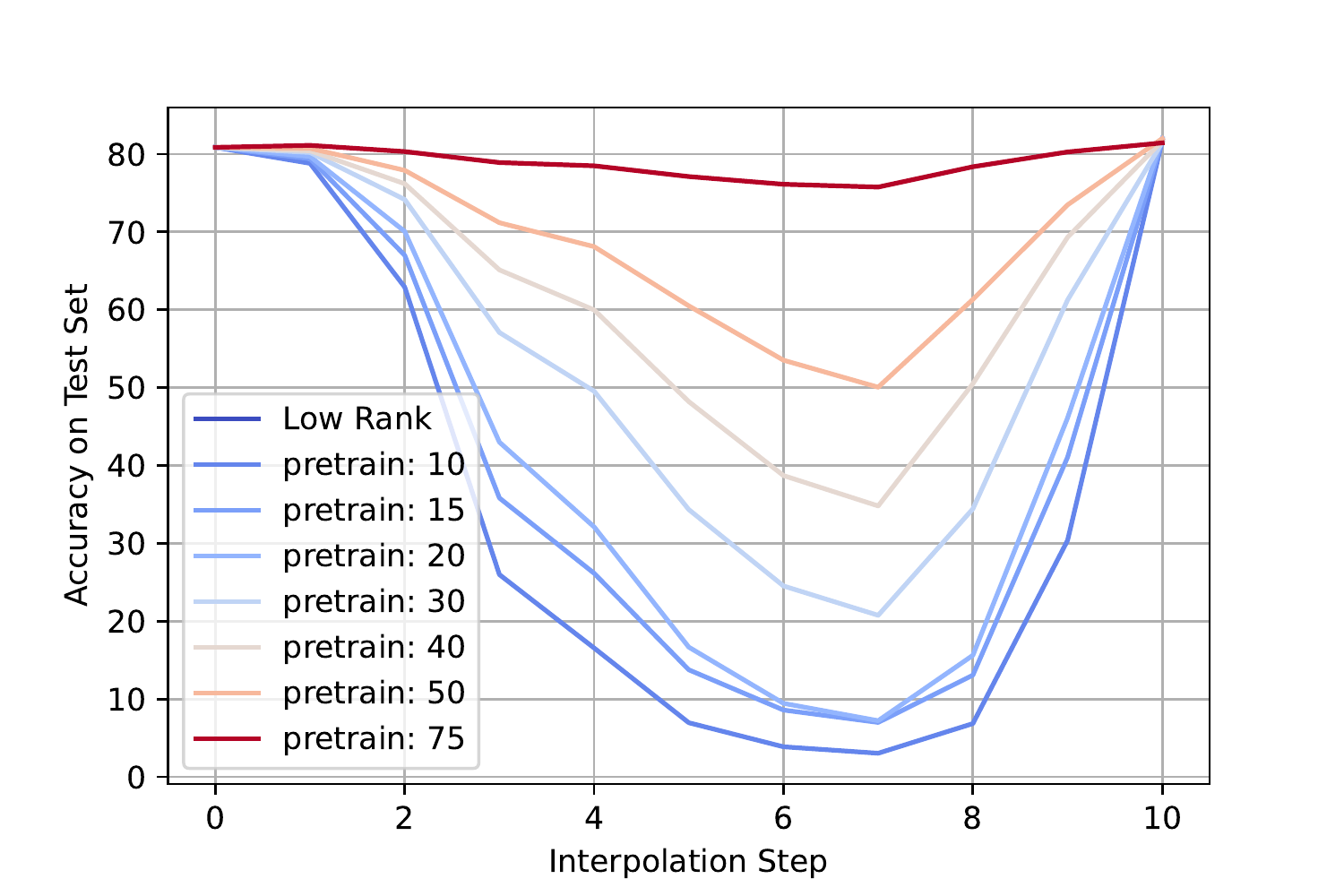}
    \caption{Comparison of interpolation of low rank and pre-trained networks for WideResNet-28 on CIFAR-100 with a rank of 30\%.}
    \label{interpolation_wrn}
\end{figure}
\begin{figure}[H]
    \centering
    \includegraphics[width=0.9\linewidth]{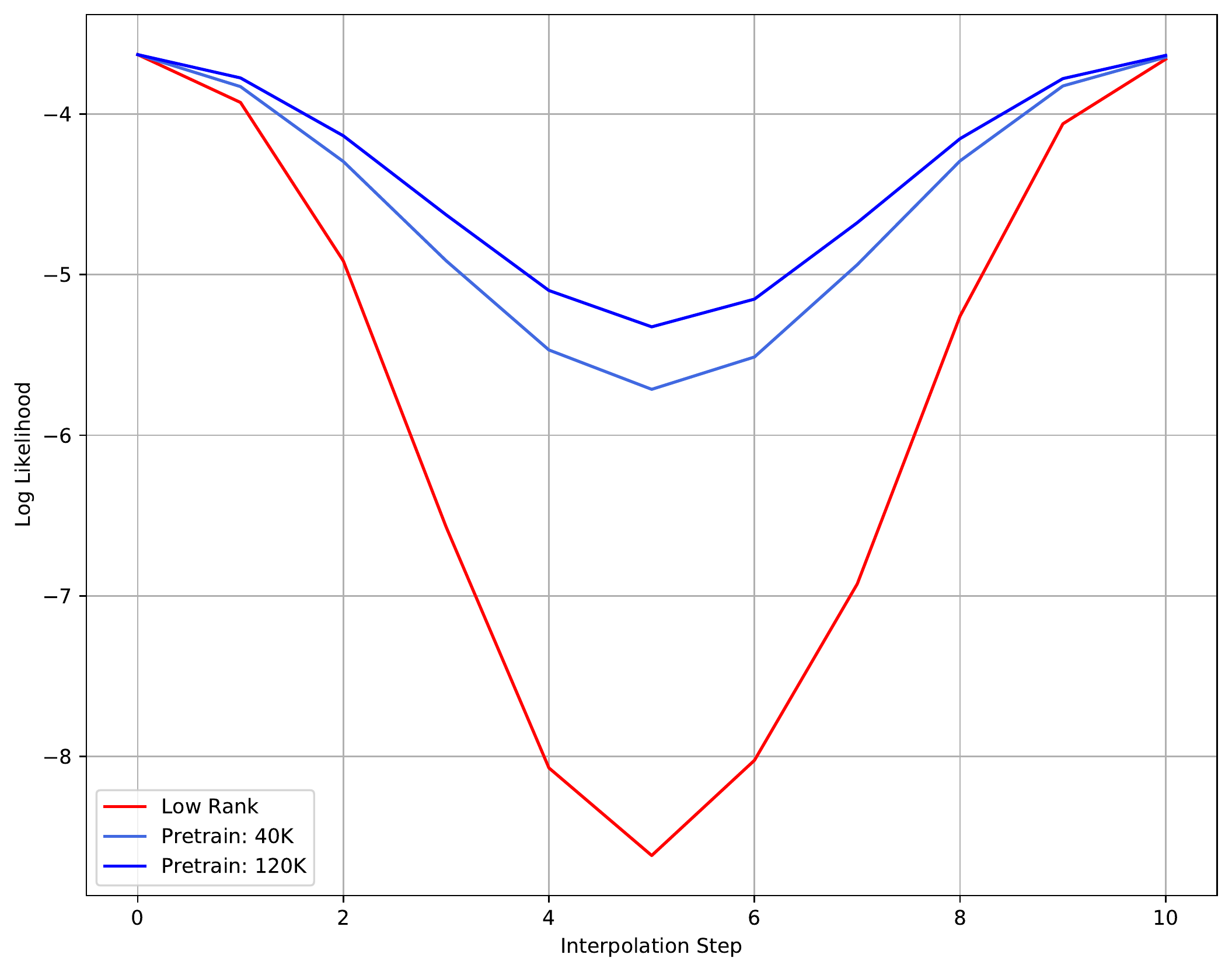}
    \caption{Comparison of interpolation of low rank and pretrained networks for transformer LM.}
    \label{interpolationLM}
\end{figure}
\end{document}